\documentclass[10pt,twocolumn,letterpaper]{article}

\usepackage{cvpr}              
\usepackage{multirow}
\usepackage{booktabs}
\usepackage[table]{xcolor}
\usepackage{caption} 
\usepackage[accsupp]{axessibility} 
\definecolor{cvprblue}{rgb}{0.21,0.49,0.74}
\usepackage[pagebackref,breaklinks,colorlinks,allcolors=cvprblue]{hyperref}

\def\paperID{15943} 
\def\confName{CVPR}
\def\confYear{2026}

\title{OpenDPR: Open-Vocabulary Change Detection via Vision-Centric Diffusion-Guided Prototype Retrieval for Remote Sensing Imagery}

\author{
Qi Guo$^{1}$ \quad
Jue Wang$^{2}$ \quad
Yinhe Liu$^{1}$\thanks{Corresponding authors.} \quad
Yanfei Zhong$^{1}$\footnotemark[1] \\
$^{1}$Wuhan University \quad
$^{2}$Beijing Institute of Technology \\
{\tt\small
\{guoqi2002, liuyinhe, zhongyanfei\}@whu.edu.cn \quad
wangjue.rs@bit.edu.cn
}
}

\begin{document}
\maketitle

\begin{abstract}
Open-vocabulary change detection (OVCD) seeks to recognize arbitrary changes of interest by enabling generalization beyond a fixed set of predefined classes. We reformulate OVCD as a two-stage pipeline: first generate class-agnostic change proposals using visual foundation models (VFMs) such as SAM and DINOv2, and then perform category identification with vision-language models (VLMs) such as CLIP. We reveal that category identification errors are the primary bottleneck of OVCD, mainly due to the limited ability of VLMs based on image-text matching to represent fine-grained land-cover categories. To address this, we propose OpenDPR, a training-free vision-centric diffusion-guided prototype retrieval framework. OpenDPR leverages diffusion models to construct diverse prototypes for target categories offline, and to perform similarity retrieval with change proposals in the visual space during inference. The secondary bottleneck lies in change localization, due to the inherent lack of change priors in VFMs. To bridge this gap, we design a spatial-to-change weakly supervised change detection module named S2C to adapt their strong spatial modeling capabilities for change localization. Integrating the pretrained S2C into OpenDPR leads to an optional weakly supervised variant named OpenDPR-W, which further improves OVCD with minimal supervision. Experimental results on four benchmark datasets demonstrate that the proposed methods achieve state-of-the-art performance under both supervision modes. Code is available at \url{https://github.com/guoqi2002/OpenDPR}.

\end{abstract}    
\vspace{-10pt}
\section{Introduction}
\label{sec:intro}

Change Detection (CD) is a fundamental task in the computer vision and remote sensing communities, which aims to localize and identify  changes in land cover by comparing multi-temporal remote sensing imagery. It has been widely applied in various domains, including environmental monitoring \cite{moharram2023monitor}, disaster response \cite{zheng2021building}, and urban planning \cite{zhu2019understanding}.

\begin{figure}[t!]
	\centering
	\setlength{\belowcaptionskip}{-6pt}
    \includegraphics[width=8.6cm, trim=0.6cm 0.6cm 0.8cm 0.6cm,clip]{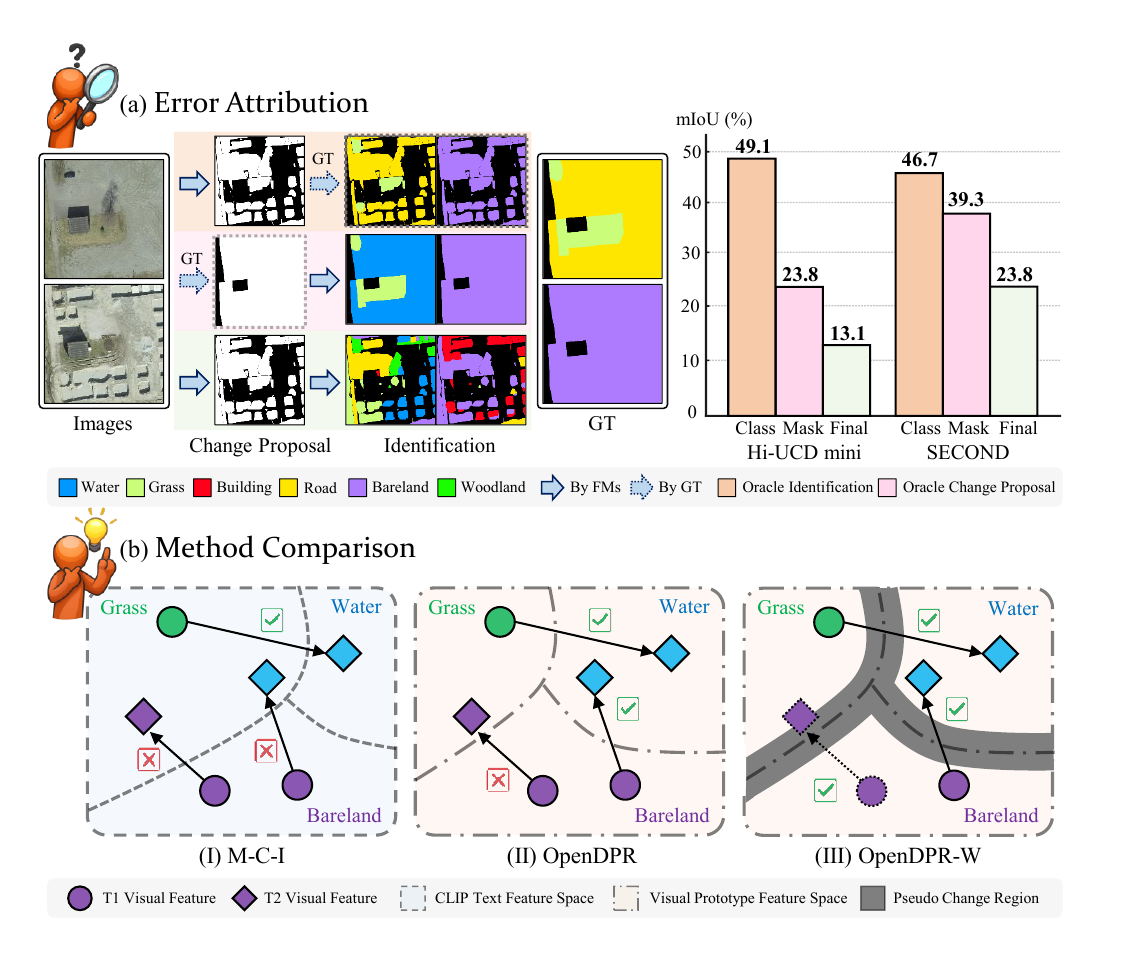}
	\caption{(a) Our analysis reveals that category identification errors are the primary bottleneck in OVCD.
(b) OpenDPR performs prototype retrieval with change proposals in the visual space to enhance category identification, while OpenDPR-W further improves change localization with additional weak supervision.
}
	\label{fig: intro}
\end{figure}

Mainstream CD approaches fall into two paradigms: binary change detection (BCD), which predicts whether a change has occurred, and semantic change detection (SCD), which further identifies the change category \cite{daudt2019multitask, yang2021asymmetric}. While differing in semantic granularity, both paradigms operate under a closed-world assumption with a fixed set of predefined categories. This constraint is particularly problematic in remote sensing scenarios characterized by diverse land cover types and scarce annotations.

To overcome the limitations of closed-set assumptions, open-vocabulary change detection (OVCD) was introduced in DynamicEarth~\cite{li2025dynamicearth}. DynamicEarth suggests that the OVCD framework should include three sequential components: mask generator, comparator, and identifier. Due to the scarcity of high-quality annotated data for large-scale pre-training \cite{peng2025lack1, tian2022lack2}, it leverages off-the-shelf visual foundation models (VFMs) and vision-language models (VLMs) to construct two training-free frameworks:
(1) M-C-I: first generates class-agnostic masks, then compares features to detect changes, and finally identifies the changed proposals;  
(2) I-M-C: first detects objects of the target category, then refines their masks, and finally compares them to detect changes.  
Since the I-M-C framework tends to accumulate errors in complex scenes, our work builds upon the more robust M-C-I framework to further advance OVCD.

We further decouple the original M-C-I framework into a two-stage pipeline: First, class-agnostic change mask proposals are generated using a combination of a mask generator and a bi-temporal comparator; Second, each proposed bi-temporal region is passed into an identifier for open-vocabulary recognition and cross-temporal comparison. 
To better diagnose the performance bottlenecks of OVCD, we design two controlled experiments to attribute errors more precisely:  
(1) \textbf{Oracle Identification}: class-agnostic change masks are generated by a mask generator and a comparator, and their semantic categories are then assigned by matching each mask with the bi-temporal land-cover ground truth via overlap-based majority voting.
(2) \textbf{Oracle Change Proposal}: ground-truth change masks are directly used as proposals, and the open-vocabulary identifier is applied to recognize their semantic categories.  
As shown in Figure~\ref{fig: intro}(a), results on Hi-UCD mini reveal that oracle identification achieves an mIoU of 49.1\%, while oracle change proposal achieves only 23.8\%, with error accumulation degrading the final OVCD performance to only 13.1\%. Similar conclusions are observed on the SECOND dataset.

Our analysis reveals that category identification errors, rather than change localization errors, are the primary bottleneck in OVCD performance. 
Specifically, open-vocabulary identifiers pre-trained on natural images such as CLIP exhibit a domain gap in remote sensing scenarios. Moreover, CLIP variants fine-tuned on remote sensing imagery \cite{liu2024remoteclip} may perform even worse due to overfitting on limited domain-specific data \cite{li2025segearth}. These observations suggest that the fundamental issue lies in the fact that textual features can only convey highly condensed global semantics and are often ambiguous for fine-grained land-cover categories, making the image-text matching paradigm of VLMs poorly suited for complex remote sensing scenarios.
 As shown in Figure~\ref{fig: intro}(b), to address this issue,  we propose a training-free framework named \textbf{OpenDPR}, which adopts vision-centric \textbf{D}iffusion-guided \textbf{P}rototype \textbf{R}etrieval to address OVCD.
Specifically, a diffusion model is employed to generate semantically diverse images for the target categories, and these images are subsequently used offline to construct a set of visual prototypes.  During inference, each change proposal retrieves its most relevant prototype based on similarity, enabling category identification entirely within the visual space.

Besides, inaccurate localization of change regions serves as a secondary bottleneck of OVCD. This issue mainly arises from the inherent lack of bi-temporal change priors in VFMs such as SAM and DINOv2, making it difficult to suppress non-semantic changes under unsupervised settings, thus leading to numerous false positives. To mitigate this problem, leveraging the strong spatial modeling capabilities of VFMs, we design a \textbf{S}patial-to-\textbf{C}hange weakly supervised change detection module named S2C to adapt this spatial awareness for binary change localization.
Specifically, given only image-level change labels for the target classes, S2C leverages VFMs to refine CAM-based pseudo labels, thereby enhancing change localization through weakly supervised pre-training.
Integrating this plug-and-play module into the training-free OpenDPR yields a weakly supervised variant named OpenDPR-W, which significantly boosts OVCD performance with minimal supervision cost.

To summarize, our contributions are as follows:

\begin{itemize}
    \item We reformulate the OVCD task as a two-stage pipeline consisting of change localization and category identification. Our analysis reveals that category identification errors are the primary performance bottleneck in OVCD.

    \item We propose a training-free vision-centric diffusion-guided prototype retrieval framework named OpenDPR, which performs prototype retrieval for change proposals within the visual space, thereby avoiding the limitations of the image-text matching paradigm used by VLMs in representing fine-grained land-cover categories.
    
    \item We further propose a weakly supervised variant named OpenDPR-W by integrating a plug-and-play weakly supervised change detection module named S2C, which leverages spatial priors from VFMs to enhance binary change localization through pre-training.

    \item Extensive experiments demonstrate that the proposed methods significantly outperform existing unsupervised and weakly supervised approaches on diverse datasets.
\end{itemize}
\section{Related Work}
\label{sec:relatedwork}

\subsection{Deep Learning-Based Change Detection}

Conventional CD methods are primarily divided into BCD and SCD. BCD focuses on locating where changes occur, typically adopting deep Siamese networks with dual encoders and a binary decoder to capture temporal differences \cite{wang2023SCPFCD, chen2024changemamba, Zang_2025_CVPR}.  
In contrast, SCD further identifies the
more challenging question of what the changes are, and can be approached via post-classification comparison, direct “from-to” prediction, or multi-task Siamese networks. Among these, multi-task Siamese networks have become the mainstream paradigm due to their ability to exploit shared feature representations between the two related tasks \cite{ding2024scannet,wang2024siamcontrast, guo2025SC2ASCD}.  
Nonetheless, both BCD and SCD approaches operate under a closed-set assumption, limiting their applicability to novel categories. Therefore, OVCD has recently been proposed by DynamicEarth \cite{li2025dynamicearth} as a more generalizable task for open-world applications.

\subsection{Foundation Models}

In recent years, foundation models have achieved remarkable success across a wide range of vision tasks. VFMs focus on extracting high-quality image representations, while VLMs align vision and language semantics by jointly learning from large-scale image-text pairs. 
Specifically, SAM~\cite{kirillov2023sam} and SAM2~\cite{ravi2024sam2} are trained on extensive image-mask pairs and support efficient interactive segmentation using prompts. DINO~\cite{caron2021dino}, DINOv2~\cite{oquab2023dinov2}, and DINOv3~\cite{simeoni2025dinov3} adopt self-supervised learning with self-distillation to learn spatially structured visual representations.
CLIP \cite{radford2021clip} aligns images and text in a shared embedding space through contrastive learning on large-scale image-text data, thus supporting open-vocabulary zero-shot recognition. APE \cite{shen2024ape} performs foundational vision tasks using an instance-level region-sentence interaction and matching paradigm, enabling unified support for visual tasks within a single model.
These foundation models play a critical role in open-world perception tasks, offering strong generalization to unseen categories and reducing reliance on dense supervision.

\subsection{Open-Vocabulary Semantic Segmentation}

Open-Vocabulary Semantic Segmentation (OVSS) aims to overcome the limitations of closed-set segmentation by enabling models to recognize and segment novel categories that are absent from the training set \cite{wu2024towardsovss, zhu2024ovsssurvey}.  Recent advancements in VLMs have significantly boosted progress in this area. Existing OVSS methods fall into two main categories: training-based and training-free. Training-based methods rely on fully supervised learning with densely annotated datasets such as COCO-Stuff \cite{caesar2018coco}, or adopt weakly supervised contrastive learning on large-scale image-text pairs \cite{xu2023OVSegmentor, cha2023TCL}.
In contrast, training-free methods improve the localization ability of CLIP through architectural modifications \cite{zhou2022maskclip, lan2024clearclip} or by incorporating auxiliary supervision from foundation models \cite{barsellotti2024freeda, zhao2025ovss}.
Compared to OVSS, OVCD remains less explored due to the additional bi-temporal comparisons and limited annotated data in the CD task.
\section{Method}

\begin{figure*}[ht]
	\centering
	\includegraphics[width=17.6cm, trim=0.8cm 0.8cm 0.8cm 0.8cm,clip]{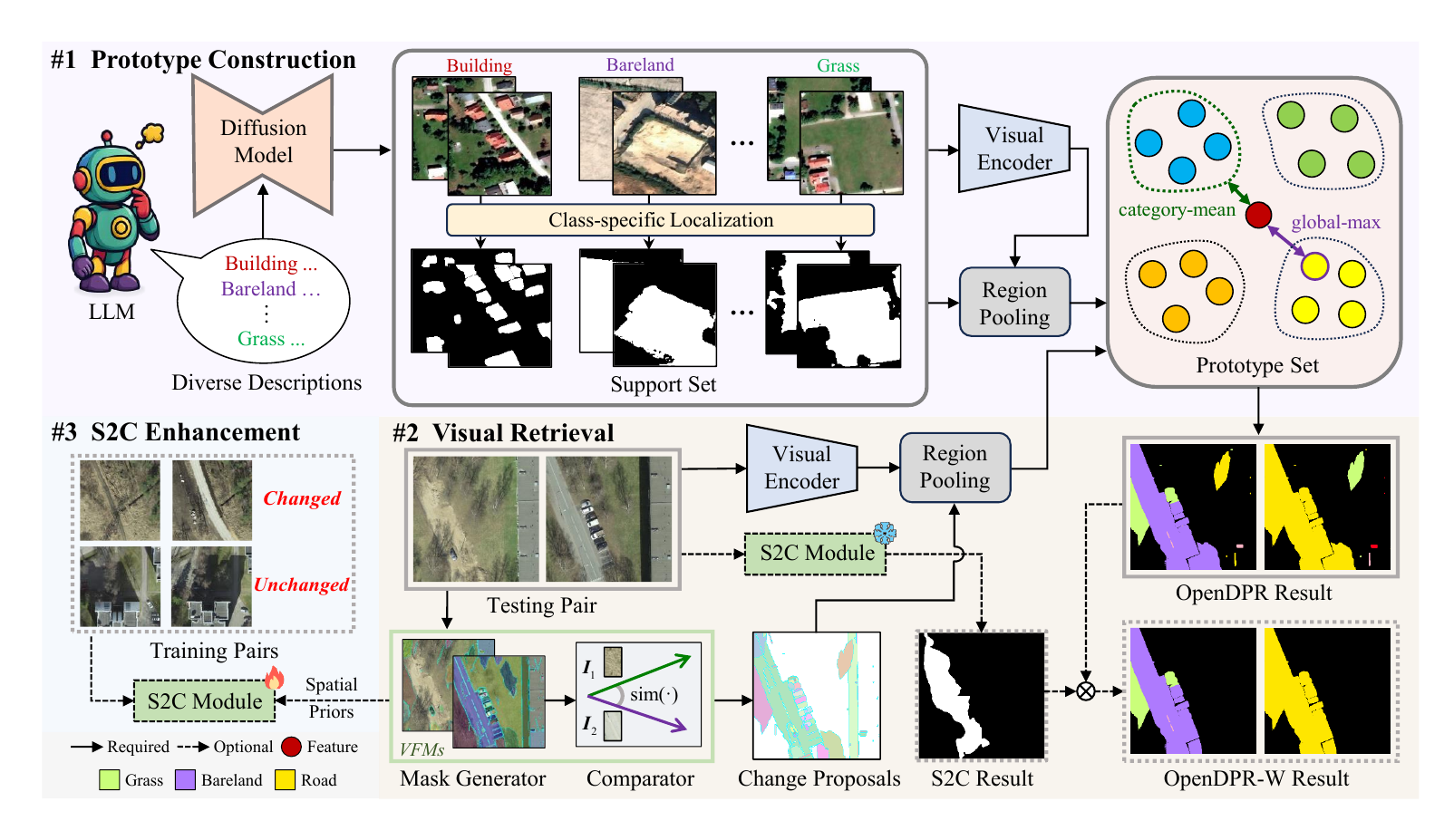}
	\caption{Overview of the proposed training-free OpenDPR and its weakly supervised variant OpenDPR-W. OpenDPR constructs a semantically diverse prototype set offline and identifies categories by retrieving change proposals against prototypes in the visual space during inference. OpenDPR-W further improves change localization by integrating the weakly supervised S2C module.
    }
	\label{fig: overview}
\end{figure*}

\subsection{Overview}

The proposed methods are illustrated in Figure~\ref{fig: overview}. The training-free OpenDPR consists of two vision-centric stages: diverse prototype construction and similarity-driven visual retrieval.
\noindent\textbf{(1) Prototype Construction.}
A large language model (LLM) is used to prompt a diffusion model to synthesize semantically diverse images of the target classes, followed by class-specific localization. Features are extracted from the synthesized images, then region-pooled within the localized class-specific regions, and clustered to obtain visual prototypes.
\noindent\textbf{(2) Visual Retrieval.}
For a test image pair, a VFM-based mask generator and comparator first produce class-agnostic change proposals. Features are then extracted from the pair, region-pooled within each candidate region, and retrieved from the prototypes based on similarity to assign change categories.
Additionally, OpenDPR-W further enhances OVCD by integrating a weakly supervised module named S2C. This variant includes an optional additional pre-training stage.
\noindent\textbf{(3) S2C Enhancement.}
S2C leverages the spatial modeling capabilities of VFMs to refine CAM-based change pseudo-labels for weakly supervised pre-training, thereby improving binary change localization.

\subsection{Diffusion-Guided Prototype Retrieval}

\subsubsection{Diverse Prototype Construction}

As shown in Figure \ref{fig: overview}, we leverage LLM to generate rich textual descriptions for each target land-cover category, thereby enhancing the diversity of images synthesized by a diffusion model tailored for remote sensing imagery.
Specifically, given a category \( c \in \mathcal{C} \) and an approximate geographic location, the prompt is formulated as: \textit{Generate \( N_1 \) diverse and detailed descriptions of [\( c \)] in remote sensing imagery. Each description should begin with ‘a satellite image of [\( c \)] in [location]’ and describe its appearance, texture, and surrounding context.}
For each description, $ N_2$ synthetic images are generated, resulting in a support set for category $c$ denoted as \( \mathcal{I}_c = \{ I_{c,i} \}_{i=1}^{N_1 \times N_2} \).

In remote-sensing imagery, target land-cover regions often constitute only a small portion of the scene, and their distributions are highly varied and complex. To prevent background semantics from leaking into class prototypes, we adopt APE \cite{shen2024ape} for class-specific localization, which generates pixel-level masks given class prompts and shows strong generalization on optical remote sensing imagery \cite{li2024semicd}.
Specifically, for each image $I_{c,i} \in \mathcal{I}_c$, a corresponding pixel-level mask $M_{c,i}$ is generated by APE based on the given class name $c$. The resulting set of masks is denoted as $\mathcal{M}_c = \{ M_{c,i} \}_{i=1}^{N_1 \times N_2}$. The complete support set for category $c$ is constructed as a collection of paired samples: $\mathcal{S}_c = (\mathcal{I}_c, \mathcal{M}_c)$. The full support set is defined as $\mathcal{S} = \{ \mathcal{S}_1, \mathcal{S}_2, \dots, \mathcal{S}_C \}$, providing category-specific, semantically aligned, and diverse supervision.

To construct discriminative and diverse visual prototypes for each land cover category, feature maps are extracted for each image $I_{c,i} \in \mathcal{I}_c$ using DINOv2 \cite{oquab2023dinov2}, which showcases good localization and semantic matching capabilities. The resulting feature map is denoted as $F_{c,i} \in \mathbb{R}^{H/p \times W/p \times d}$, where $H$ and $W$ denote the height and width of the original image, $d$ is the feature dimension, and $p$ is the patch size of the feature extractor. Each feature map \( F_{c,i} \) is upsampled to the original image resolution, and average pooling is performed over the localization mask \( M_{c,i} \) to aggregate visual features within the target region:

\begin{equation}
z_{c,i} = \frac{\sum_{h=0}^{H} \sum_{w=0}^{W} M_{c,i}(h,w)\,{F}_{c,i}(h,w)}{\sum_{h=0}^{H} \sum_{w=0}^{W} M_{c,i}(h,w)},
\label{eq:masked_pool}
\end{equation}
where $(h,w)$ denote the spatial coordinates. The set of region-level features for category $c$ is denoted as $\mathcal{Z}_c = \{ z_{c,i} \}_{i=1}^{N_1 \times N_2}$. To capture the intra-class diversity within $\mathcal{Z}_c$, clustering is performed over the feature set. Each cluster centroid serves as an implicit attribute, representing a subset of land covers with consistent semantic characteristics:

\begin{equation}
\mathcal{P}_c = \{ p_{c,k} \}_{k=1}^{K} = \text{KMeans}(\mathcal{Z}_c, K),
\label{eq:prototypes_kmeans}
\end{equation}
where $K$ denotes the number of centroids, and K-means clustering algorithm \cite{lloyd1982kmeans} is adopted for simplicity. The final prototype set across all categories is defined as $\mathcal{P} = \{ \mathcal{P}_1, \mathcal{P}_2, \dots, \mathcal{P}_C \}$. The above prototype construction process is efficient and scalable, since it is performed entirely offline and can be reused across downstream scenarios.

\subsubsection{Similarity-Driven Visual Retrieval}

During inference, given a pair of bi-temporal images $\mathit{I}_t = \{I_1, I_2\} \in \mathcal{D}_{\mathrm{test}}$, we first use a VFM-based mask extractor and comparator to generate class-agnostic change proposals. For each proposed instance, we extract visual features and retrieve them from the prototype set $\mathcal{P}$ to derive the final OVCD predictions.

Specifically, SAM~\cite{kirillov2023sam} is applied to $\mathit{I}_t$ to obtain initial mask sets $\mathcal{M}_1$ and $\mathcal{M}_2$. To eliminate redundant detections caused by unchanged objects, duplicate masks are removed by applying non-maximum suppression (NMS) to the union of $\mathcal{M}_1$ and $\mathcal{M}_2$, resulting in the final land-cover instance masks $\mathcal{M} = \mathrm{NMS}(\mathcal{M}_1 \cup \mathcal{M}_2)$.
Subsequently, the feature maps of $\mathit{I}_t$ are individually extracted using DINOv2~\cite{oquab2023dinov2}, interpolated to match the spatial resolution of $\mathcal{M}$, and aggregated through masked average pooling as defined in equation \ref{eq:masked_pool} to obtain the masked feature set $\mathcal{Z}_t = \{(z_1^{(m)}, z_2^{(m)})\}_{m=1}^{|\mathcal{M}|}$.
The change score is derived from the negative cosine similarity between paired features. Masks with scores above the threshold $\alpha$ are retained as predicted change proposals $\hat{\mathcal{M}}$.
To identify the land-cover category for each predicted change proposal \(M_i \in \hat{\mathcal{M}}\), cosine similarity is computed between the masked feature \(z_i \in \mathcal{Z}_{t}\) and prototype \(p_{c,k} \in \mathcal{P}\):

\begin{equation}
\text{sim}\bigl(z_{i}, p_{c,k}) 
= \frac{z_{i} \cdot p_{c,k}}{\|z_{i}\|_2\, \|p_{c,k}\|_2},
\end{equation}
where $i \in \{1, 2\}$. We propose two effective and intuitive strategies for retrieving class labels for each change proposal. In the \textit{category-mean} strategy, each change proposal is assigned to the class whose prototypes yield the highest average similarity:

\begin{equation}
\hat{c}_i = \arg\max_{c} \; \frac{1}{K} \sum_{k=1}^{K} \text{sim}(z_i, p_{c,k}),
\end{equation}

Alternatively, in the \textit{global-max} strategy, each mask is assigned to the class that is associated with the single most similar prototype across all categories:

\begin{equation}
\hat{c}_i = \arg\max_{c,k} \;\text{sim}(z_i, p_{c,k}),
\end{equation}

A detailed comparison of the two strategies is presented in Section~\ref{sssec:OpenDPR-components}.

\subsection{Weak Supervision for OpenDPR}

\begin{figure}[!t]
	\centering
	\setlength{\belowcaptionskip}{-6pt}
    \includegraphics[width=8.5cm, trim=0.5cm 0.8cm 0.8cm 0.5cm,clip]{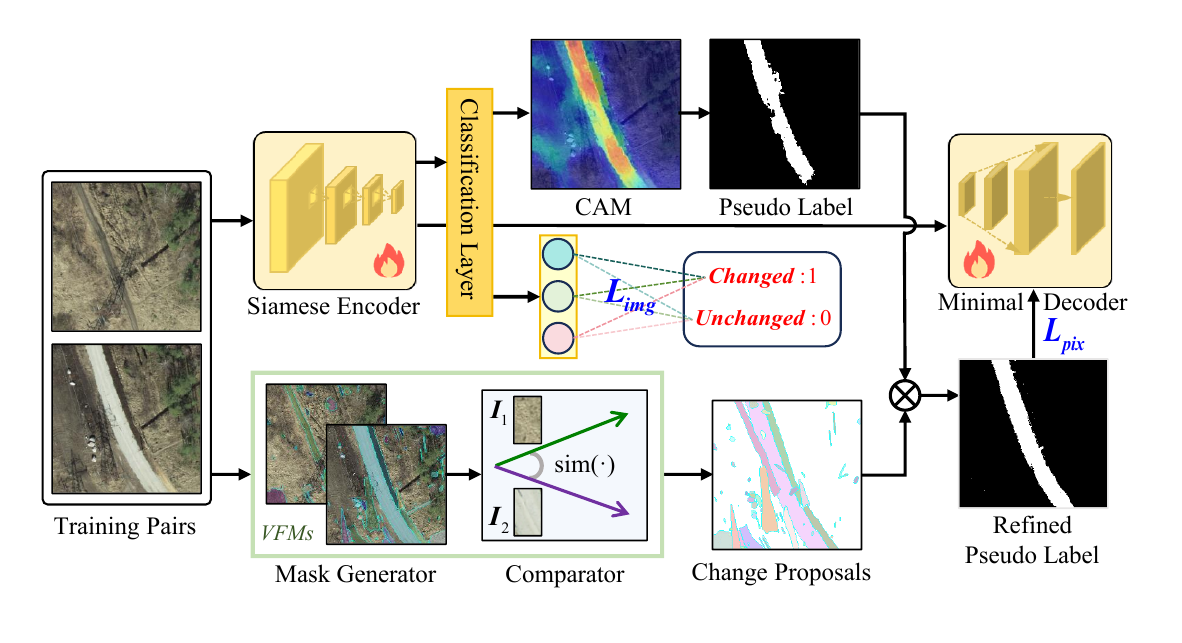}
	\caption{Overview of the proposed S2C, which refines CAM-based change pseudo-labels using spatial priors from VFMs and adapts them to binary change localization through pre-training.}
	\label{fig: S2C}
\end{figure}

\subsubsection{S2C Pre-training}

As shown in Figure \ref{fig: S2C}, S2C retains only the essential modules of the standard weakly supervised change detection paradigm. Given a pair of bi-temporal images denoted as $I_t = \{I_1, I_2 \} \in \mathcal{D}_{\mathrm{train}}$, the corresponding image-level change label is $Y_{\text{cls}} \in \{0, 1\}$. Both images are fed into a shared-weight Siamese encoder to extract multi-scale hierarchical features. The final layer feature maps from each branch, denoted as \( F_1 \) and \( F_2 \), are selected and fused via element-wise absolute difference to form the change feature \( F_f = |F_1 - F_2| \).
Global average pooling is then applied to \( F_f \), followed by a \( 1 \times 1 \) convolution to generate the predicted logit \( p_{\text{img}} \), which is used to compute the image-level binary cross-entropy loss 
\(\mathcal{L}_{\text{img}}=\ell_{\mathrm{bce}}\big(Y_{\text{cls}},\,p_{\text{img}}\big)\).


 Since image-level change labels cannot supervise pixel-level localization, class activation map (CAM) is adopted to generate pseudo labels for dense supervision \cite{yang2025excel}. CAM is computed by applying a class-specific weighted summation over the change feature map \(F_f\) using the classifier weights \(W_{\text{cls}}\). The resulting CAM is min-max normalized and binarized using a threshold \(\beta\), yielding the pixel-level pseudo label \(\hat{Y}_{\text{pix}}\). To generate pixel-level predictions, the change feature \(F_f\) is fed into a lightweight segmentation head comprising an MLP, a dropout layer, and two \(1 \times 1\) convolutions to obtain \(p_{\text{pix}}\), which is used for computing the pixel-level binary cross-entropy loss \(\mathcal{L}_{\text{pix}}=\ell_{\mathrm{bce}}(\hat{Y}_{\text{pix}},\,p_{\text{pix}})\). The final loss for S2C pre-training is defined as:
\begin{equation}
\mathcal{L} = \mathcal{L}_{\text{img}} + \lambda \mathcal{L}_{\text{pix}},
\label{eq:s2c_total_loss}
\end{equation}
where $\lambda$ is a weight coefficient. Since CAM typically highlights only the most discriminative change regions, the resulting noisy pseudo labels potentially lead to suboptimal supervision. Therefore, S2C further incorporates the strong spatial capability of VFMs to improve the quality of pseudo labels \(\hat{Y}_{\text{pix}}\) derived from ${\text{CAM}}$ for pre-training.
Specifically, consistent with the unsupervised change localization pipeline in OpenDPR, SAM~\cite{kirillov2023sam} and DINOv2~\cite{oquab2023dinov2} are first employed to extract class-agnostic change proposals \(\hat{\mathcal{M}}\) from the train pair \(I_t = \{I_1, I_2\} \in \mathcal{D}_{\mathrm{train}}\).
For each change proposal \(M_i \in \hat{\mathcal{M}}\), the overlap ratio \(r_i\) with the initial coarse pseudo label \(\hat{Y}_{\text{pix}}\) is computed as:

\begin{equation}
r_i=\frac{|M_i\cap\hat{Y}_{\text{pix}}|}{|M_i|},
\label{eq:overlap}
\end{equation}
where \(|\cdot|\) denotes the pixel count. Change proposals satisfying \(r_i > \gamma_1\) are fused to generate a refined pseudo label \(\hat{Y}_{\text{pix}}^{\text{ref}}\) with more accurate and spatially aligned change boundaries, which is then used to compute \(\mathcal{L}_{\text{pix}}\).

\subsubsection{Inference-time Enhancement}

S2C adapts the spatial priors of VFMs to binary change localization through pre-training. During inference, S2C is first employed to predict the binary change region \(\mathcal{R}\) from the test image pair \(I_t = \{I_1, I_2\} \in \mathcal{D}_{\mathrm{test}}\). For each change proposal \(M_i \in \hat{\mathcal{M}}\) predicted by OpenDPR, its overlap ratio \(r_i\) with the change region \(\mathcal{R}\) is computed according to equation~\ref{eq:overlap}. Change proposals with overlap ratios exceeding a predefined threshold $\gamma_2$  are retained and aggregated to form the final OVCD result of OpenDPR-W.

\section{Experiments}
\label{sec:Experiments}

\subsection{Datasets and Evaluation Metrics}

To fully evaluate the proposed methods, experiments are conducted on four diverse datasets: LEVIR-CD~\cite{chen2020levircd} and WHU-CD~\cite{ji2018whucd} for building change detection, and Hi-UCD mini~\cite{tian2020HiUCDmini} and SECOND~\cite{yang2021asymmetric} for semantic change detection. For OpenDPR, inference is performed in a training-free manner using only the test set. For OpenDPR-W, weakly supervised pre-training is first conducted on the training set, and the model achieving the best F1 score on the validation set is used for binary change localization on the test set. These results are fused with the training-free outputs from OpenDPR. (mean) IoU and (mean) F1 are reported as evaluation metrics. Notably, as each image pair in the manually curated SECOND dataset inherently contains change, only OpenDPR is evaluated on this dataset. More details are provided in the Supplementary Material.

\subsection{Implementation Details}

In OpenDPR, GPT-4 is used to generate \(N = 20\) diverse descriptions for each land cover category, and DiffusionSat~\cite{khanna2024diffusionsat} is employed to synthesize \(M = 5\) support images per description. The number of centroids \(K\) is set to 10 for the SECOND dataset and 5 for all others. The comparator threshold \(\alpha\) follows the setting in~\cite{li2025dynamicearth} across all datasets.
In OpenDPR-W, the encoder of S2C is implemented using MiT-b1, as its hierarchical design is better suited for segmentation tasks~\cite{liu2025acwcd}. A batch size of 8 is used for pre-training over 10{,}000 iterations, with each input image pair resized to \(256 \times 256\). The loss balance coefficient $\lambda$ is set to 1, the CAM binarization threshold \(\beta\) is set to 0.5, and the mask filtering thresholds \(\gamma_1\) and \(\gamma_2\) are both set to 0.05.

\subsection{Comparison with SOTA}

The proposed methods target OVCD under training-free and image-level weakly supervised settings. For binary building change detection, we compare against unsupervised methods including CVA~\cite{bovolo2006CVA}, PCA-KM~\cite{celik2009PCAKM}, CNN-CD~\cite{el2016CNN-CD}, DSFA~\cite{du2019DSFA}, DCVA~\cite{saha2019DCVA}, GMCD~\cite{tang2021GMCD}, DINOv2+CVA~\cite{zheng2024segmentanychange}, AnyChange~\cite{zheng2024segmentanychange}, SCM~\cite{tan2024scm}, and M-C-I~\cite{li2025dynamicearth}, as well as weakly supervised methods including FCD-GAN~\cite{wu2023fcd-gan}, BGMix~\cite{huang2023bgmix}, CS-WSCDNet~\cite{wang2023cswscdnet}, MS-Former~\cite{li2024msformer}, TransWCD~\cite{zhao2025transwcd}, and ACWCD~\cite{liu2025acwcd}. For multi-class semantic change detection, we conduct a comprehensive comparison with existing OVCD methods I-M-C~\cite{li2025dynamicearth} and M-C-I~\cite{li2025dynamicearth}.

\subsubsection{Results on Building Change Detection}

Tables~\ref{tab:BCD-free} and~\ref{tab:BCD-weak} summarize the performance on the LEVIR-CD and WHU-CD datasets. Under the unsupervised setting, OpenDPR consistently outperforms existing approaches, achieving IoU improvements of 8.2\% and 13.7\% over the previous best M-C-I, thereby demonstrating superior generalization compared to CLIP-based category identification in remote sensing scenarios. With image-level weak supervision, the proposed S2C surpasses all existing weakly supervised methods while maintaining the smallest trainable parameters. By integrating S2C’s change localization with OpenDPR’s category identification capabilities, OpenDPR-W further improves IoU by 3.3\% and 14.0\% over training-free OpenDPR, and by 5.9\% and 4.6\% over the previous best weakly supervised method ACWCD.

\begin{table}[ht]
  \centering
  \fontsize{9.3pt}{10pt}\selectfont
  \setlength{\tabcolsep}{8pt}
  \renewcommand{\arraystretch}{1.2}
  \caption{Comparison of unsupervised methods on building change detection datasets. “—” denotes data missing. Best and second-best scores are marked in \textbf{bold} and in \underline{underline}, respectively.}
  \label{tab:BCD-free}

  \begin{tabular}{l|cc|cc} 
    \hline
    \multirow{2}[2]{*}{Method} & \multicolumn{2}{c|}{LEVIR-CD} & \multicolumn{2}{c}{WHU-CD} \\
          & IoU & F1 & IoU & F1 \\
    \hline
    CVA~\cite{bovolo2006CVA}         & —    & 12.2 & —    & —   \\
    PCA-KM~\cite{celik2009PCAKM}     & 4.8  & 9.1  & 5.4  & 10.2 \\
    CNN-CD~\cite{el2016CNN-CD}       & 7.0  & 13.1 & 4.9  & 9.4  \\
    DSFA~\cite{du2019DSFA}           & 4.3  & 8.2  & 4.1  & 7.8  \\
    DCVA~\cite{saha2019DCVA}         & 7.6  & 14.1 & 10.9 & 19.6 \\
    GMCD~\cite{tang2021GMCD}         & 6.1  & 11.6 & 10.9 & 19.7 \\
    DINOv2+CVA~\cite{zheng2024segmentanychange}  & —    & 17.3 & —    & —   \\
    AnyChange-H~\cite{zheng2024segmentanychange} & —    & 23.0 & —    & —   \\
    SCM~\cite{tan2024scm}            & 18.8 & 31.7 & 18.6 & 31.3 \\
    M-C-I~\cite{li2025dynamicearth}  & \underline{36.6} & \underline{53.6} & \underline{40.6} & \underline{57.7} \\
    \rowcolor{gray!12}
    \textbf{OpenDPR (Ours)} & \textbf{44.8} & \textbf{61.9} & \textbf{54.3} & \textbf{70.4} \\
    \hline
  \end{tabular}
\end{table}

\begin{table}[ht]
  \centering
  \fontsize{9.3pt}{10pt}\selectfont
  \setlength{\tabcolsep}{5.2pt}
  \renewcommand{\arraystretch}{1.2}
  \caption{Comparison of weakly supervised methods on building change detection datasets. “Params” denotes trainable parameters.}
  \label{tab:BCD-weak}

  \begin{tabular}{l|cc|cc|c} 
    \hline
    \multirow{2}[2]{*}{Method} & \multicolumn{2}{c|}{LEVIR-CD} & \multicolumn{2}{c|}{WHU-CD} & \multirow{2}[2]{*}{Params} \\
          & IoU & F1 & IoU & F1 & \\
    \hline
    FCD-GAN~\cite{wu2023fcd-gan}       & 30.5 & 43.1 & 39.3 & 56.5 & 43.4M \\
    BGMix~\cite{huang2023bgmix}        & 38.1 & 52.0 & 42.7 & 59.8 & 22.1M \\
    CS-WSCDNet~\cite{wang2023cswscdnet} & 40.7 & 57.8 & 57.3 & 72.8 & 99.9M \\
    MS-Former~\cite{li2024msformer}    & 39.8 & 57.0 & 51.3 & 67.8 & 15.2M \\
    TransWCD~\cite{zhao2025transwcd}   & 42.9 & 60.1 & 52.4 & 68.7 & 17.9M \\
    ACWCD~\cite{liu2025acwcd}          & 42.2 & 59.3 & 63.7 & 77.9 & 18.1M \\
    \rowcolor{gray!12}
    \textbf{S2C (Ours)} & \underline{44.7} & \underline{61.8} & \underline{64.8} & \underline{78.7} & \textbf{13.4M} \\
    \rowcolor{gray!12}
    \textbf{OpenDPR-W (Ours)} & \textbf{48.1} & \textbf{65.0} & \textbf{68.3} & \textbf{81.2} & \textbf{13.4M} \\
    \hline
  \end{tabular}
\end{table}


\subsubsection{Results on Semantic Change Detection}

Table~\ref{tab:Hi-UCD mini} and Table~\ref{tab:SECOND-mini} present the results on the Hi-UCD mini and SECOND datasets. Both M-C-I and I-M-C were originally designed for per-class inference. We extend M-C-I to support efficient multi-class inference, where M-C-I (SP) and M-C-I (MP) denote the use of single and multiple textual prompts per category within the open-vocabulary identifier, respectively. In contrast, I-M-C cannot be directly extended, so only its per-class results are reported.

Across different inference settings, M-C-I exhibits large performance variations, primarily due to the sensitivity of its open-vocabulary identifier. In comparison, OpenDPR consistently outperforms all comparison methods by improving category identification, while OpenDPR-W further enhances performance through better change localization. More importantly, our methods perform better while significantly reducing inference time via simultaneous multi-class inference, demonstrating strong practicality for OVCD applications. For rare and complex categories such as pedestrian bridge in Hi-UCD mini, all methods fail to detect them. These failures are mainly caused by semantic biases in visual prototype construction.


\newlength\MethodCol
\setlength{\MethodCol}{30mm} 

\newlength\ColW             
\setlength{\tabcolsep}{1.1pt} 

\setlength\ColW{\dimexpr
  (\textwidth - \MethodCol - 10\arrayrulewidth - 36\tabcolsep)/18
\relax}

\newcolumntype{Q}{>{\centering\arraybackslash}m{\ColW}}
\newcolumntype{L}{>{\raggedright\arraybackslash}m{\MethodCol}}

\begin{table*}[t]
  \centering
  \caption{Comparison of OVCD methods on the semantic change detection dataset Hi-UCD mini. “—” denotes that the score is close to 0.}
  \label{tab:Hi-UCD mini}

  {\fontsize{9.2}{10.8}\selectfont
  \renewcommand{\arraystretch}{1.15}
  \aboverulesep=0pt \belowrulesep=0pt

  \begin{tabular}{@{}L
    | QQ | QQ | QQ | QQ | QQ | QQ | QQ | QQ || QQ @{}}
    \toprule
    \multirow{2}{*}{Method} &
      \multicolumn{2}{c|}{Water} &
      \multicolumn{2}{c|}{Grass} &
      \multicolumn{2}{c|}{Building} &
      \multicolumn{2}{c|}{Greenhouse} &
      \multicolumn{2}{c|}{Road} &
      \multicolumn{2}{c|}{Bridge} &
      \multicolumn{2}{c|}{Bareland} &
      \multicolumn{2}{c||}{Woodland} &
      \multicolumn{2}{c}{Average} \\
    & IoU & F1 & IoU & F1 & IoU & F1 & IoU & F1 & IoU & F1 & IoU & F1 & IoU & F1 & IoU & F1 & mIoU & mF1 \\
    \midrule

    \textcolor{gray}{\textit{\textbf{Per-class Inference}}}
      & & & & & & & & & & & & & & & & & & \\

    I-M-C~\cite{li2025dynamicearth}
      & \textbf{10.0} & \textbf{18.1} & 9.2 & 16.8 & \underline{35.1} & \underline{52.0} & — & — & 18.6 & 31.3 & — & — & — & — & — & — & 9.1 & 14.8 \\
    M-C-I~\cite{li2025dynamicearth}
      & 2.6 & 5.0 & 21.0 & 34.8 & 22.7 & 37.1 & 8.9 & 16.3 & 22.3 & 36.5 & — & — & 26.5 & 41.9 & — & — & 13.0 & 21.5 \\
    \rowcolor{gray!12} \textbf{OpenDPR (Ours)}
      & 2.9 & 5.5 & 26.6 & 42.1 & 26.5 & 41.8 & 15.1 & 26.2 & 18.2 & 30.8 & — & — & 24.2 & 39.0 & 1.7 & 3.4 & 14.4 & 23.6 \\
    \rowcolor{gray!12} \textbf{OpenDPR-W (Ours)}
      & 4.4 & 8.4 & \textbf{32.6} & \textbf{49.2} & 34.2 & 51.0 & 12.9 & 22.8 & 21.6 & 35.5 & — & — & \underline{29.0} & \underline{44.9} & \textbf{3.3} & \textbf{6.4} & 17.3 & 27.3 \\
    \midrule

    \textcolor{gray}{\textit{\textbf{Multi-class Inference}}}
      & & & & & & & & & & & & & & & & & & \\

    M-C-I (SP)
      & — & — & 21.4 & 35.2 & 21.6 & 35.5 & 11.9 & 21.3 & 21.1 & 34.8 & — & — & 3.0 & 5.9 & — & — & 9.9 & 16.6 \\
    M-C-I (MP)
      & 2.5 & 5.0 & 24.0 & 38.7 & 23.8 & 38.4 & 9.5 & 17.4 & \textbf{23.3} & \textbf{37.7} & — & — & 27.8 &43.5 & — & — & 13.9 & 22.6 \\
    \rowcolor{gray!12} \textbf{OpenDPR (Ours)}
      & \underline{5.1} & \underline{9.8} & 24.4 & 39.2 & 34.0 & 50.8 & \textbf{29.0} & \textbf{45.0} & 18.0 & 30.5 & — & — & 27.8 & 43.5 & 1.9 & 3.6 & \underline{17.5} & \underline{27.8} \\
    \rowcolor{gray!12} \textbf{OpenDPR-W (Ours)}
  & 4.7 & 8.9 & \underline{28.5} & \underline{44.3} & \textbf{41.9} & \textbf{59.0} & \underline{24.7} & \underline{39.6} & \underline{23.2} & \underline{37.6} & — & — & \textbf{31.8} & \textbf{48.2} & \underline{2.9} & \underline{5.6} & \textbf{19.7} & \textbf{30.4} \\

    \bottomrule
  \end{tabular}
  
  }
\end{table*}



\setlength{\MethodCol}{30mm}  

\setlength{\tabcolsep}{1.2pt} 

\newlength\TableWidth
\setlength\TableWidth{1\textwidth} 

\setlength\ColW{\dimexpr
  (\TableWidth - \MethodCol - 8\arrayrulewidth - 28\tabcolsep)/14
\relax}

\newcolumntype{Q}{>{\centering\arraybackslash}m{\ColW}}
\newcolumntype{L}{>{\raggedright\arraybackslash}m{\MethodCol}}

\begin{table*}[t]
  \centering
  \caption{Comparison of OVCD methods on the semantic change detection dataset SECOND. “—” denotes that the score is close to 0.}
  \label{tab:SECOND-mini}

  {\fontsize{9.2}{10.8}\selectfont
  \renewcommand{\arraystretch}{1.15}
  \aboverulesep=0pt \belowrulesep=0pt

  \begin{tabular}{@{}L | QQ | QQ | QQ | QQ | QQ | QQ || QQ @{}}
    \toprule
    \multirow{2}{*}{Method} &
      \multicolumn{2}{c|}{Water} &
      \multicolumn{2}{c|}{Ground} &
      \multicolumn{2}{c|}{Low vegetation} &
      \multicolumn{2}{c|}{Tree} &
      \multicolumn{2}{c|}{Building} &
      \multicolumn{2}{c||}{Playground} &
      \multicolumn{2}{c}{Average} \\
    & IoU & F1 & IoU & F1 & IoU & F1 & IoU & F1 & IoU & F1 & IoU & F1 & mIoU & mF1 \\
    \midrule

    \textcolor{gray}{\textit{\textbf{Per-class Inference}}} & & & & & & & & & & & & & & \\

    I-M-C~\cite{li2025dynamicearth}
      & 12.2 & 21.7 & — & — & 1.4 & 2.7 & 14.1 & 24.8 & 28.1 & 43.9 & 16.0 & 27.6 & 12.0 & 20.1 \\
    M-C-I~\cite{li2025dynamicearth}
      & \underline{14.3} & \underline{25.1} & 26.2 & 41.6 & \textbf{24.1} & \textbf{38.9} & \underline{20.3} & \underline{33.8} & 38.1 & 55.2 & 20.0 & 33.3 & 23.8 & 38.0 \\
    \rowcolor{gray!12}
    \textbf{OpenDPR (Ours)}
      & 12.0 & 21.4 & \textbf{31.2} & \textbf{47.5} & \underline{23.9} & \underline{38.6} & 19.7 & 33.0 & \textbf{43.2} & \textbf{60.4} & \underline{26.5} & \underline{42.0} & \underline{26.1} & \underline{40.5} \\
    \midrule

    \textcolor{gray}{\textit{\textbf{Multi-class Inference}}} & & & & & & & & & & & & & & \\

    M-C-I (SP)
      & 8.4 & 15.5 & 24.3 & 39.1 & 9.4 & 17.2 & 9.0 & 16.5 & 5.1 & 9.8 & — & — & 9.4 & 16.4 \\
    M-C-I (MP)
      & 10.9 & 19.6 & 25.4 & 40.5 & 22.5 & 36.7 & \underline{20.3} & \underline{33.8} & 30.8 & 47.1 & 23.3 & 37.8 & 22.2 & 35.9 \\
    \rowcolor{gray!12}
    \textbf{OpenDPR (Ours)}
      & \textbf{17.5} & \textbf{29.8} & \underline{30.2} & \underline{46.4} & 23.0 & 37.4 & \textbf{20.9} & \textbf{34.5} & \underline{42.4} & \underline{59.5} & \textbf{37.3} & \textbf{54.3} & \textbf{28.6} & \textbf{43.7} \\
    \bottomrule
  \end{tabular}
  }
\end{table*}

\subsubsection{Visualization}

\setlength{\belowcaptionskip}{-6pt}
\begin{figure}[ht]
	\centering
	\includegraphics[width=8.5cm, trim=0.5cm 0.9cm 0.5cm 0.7cm,clip]{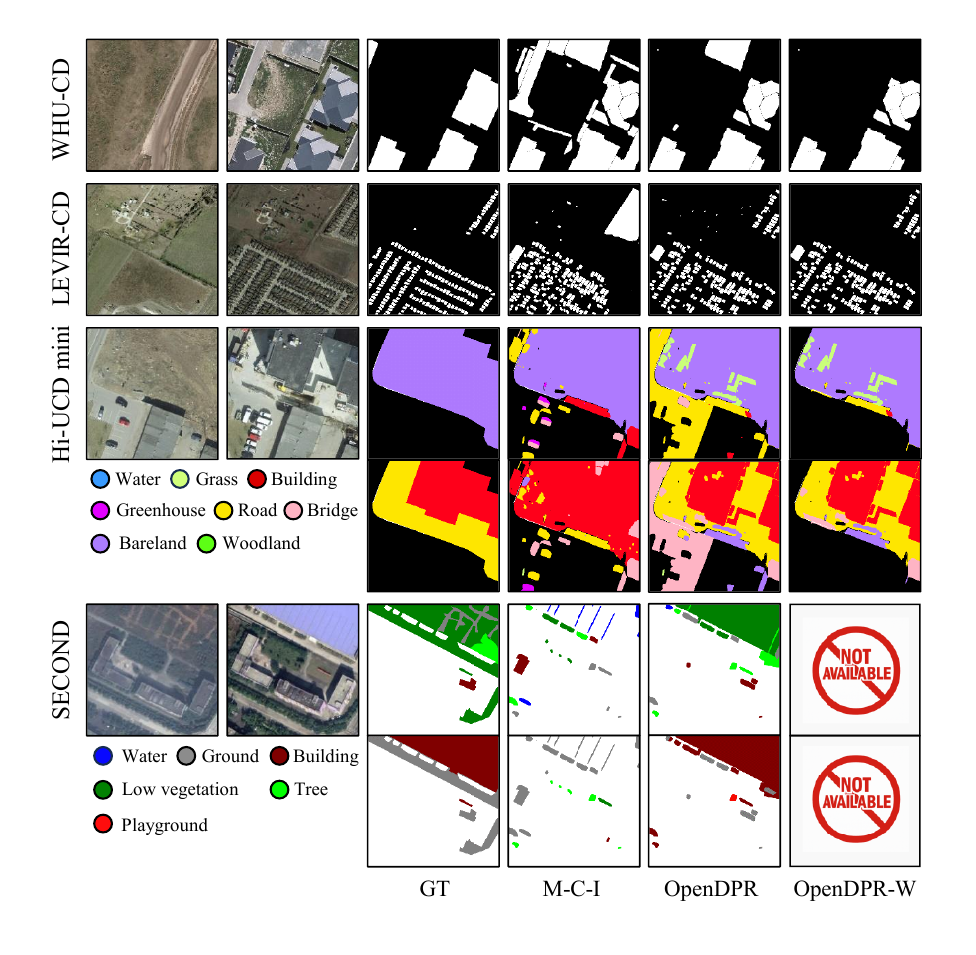}
	\caption{Qualitative evaluation of the proposed methods on four change detection datasets.}
	\label{fig: viz}
\end{figure}

Figure~\ref{fig: viz} presents qualitative comparisons on four change detection datasets. Compared to M-C-I, OpenDPR demonstrates markedly improved land cover recognition, especially in complex scenes such as Hi-UCD mini and SECOND. Furthermore, the weakly supervised OpenDPR-W effectively reduces false alarms in change localization, highlighting the potential of incorporating additional supervision to enhance OVCD performance. More visualizations are provided in the Supplementary Material.

\subsection{Ablation Studies and Analyses}

\subsubsection{Different Components in OpenDPR}
\label{sssec:OpenDPR-components}

Table~\ref{tab:ablation1} reports the impact of each component of the proposed training-free OpenDPR across four change detection datasets.
For prototype construction, we compare prompts generated by LLM with naive prompts such as \textit{“a satellite image of [class] in [location].”} Results show that LLM-generated prompts generally yield better performance, as they provide richer semantic information for synthesizing support images.
For the visual encoder, we evaluate three representative foundation models: CLIP, DINO, and DINOv2. Among them, DINOv2 exhibits the best overall performance, highlighting the advantage of stronger visual representations in modeling remote sensing land covers.
For the matching strategy, we compare category-mean and global-max strategies. The global-max strategy generally performs better by capturing the strongest semantic alignment between change proposals and support prototypes, effectively avoiding the dilution from feature averaging in diverse remote sensing scenes.

\begin{table}[ht]
  \centering
  \fontsize{9.3pt}{10pt}\selectfont
  \setlength{\tabcolsep}{5.2pt}
  \renewcommand{\arraystretch}{1.2}
  \caption{Ablation study on (mean) IoU of different components in OpenDPR. \(\mathcal{L}\), \(\mathcal{W}\), \(\mathcal{H}\), and \(\mathcal{S}\) denote the LEVIR-CD, WHU-CD, Hi-UCD mini, and SECOND datasets, respectively.}
  \begin{tabular}{c c c | c c c c}
    \hline  
    Proto & Encoder & Match & \(\mathcal{L}\) & \(\mathcal{W}\) & \(\mathcal{H}\) & \(\mathcal{S}\) \\
    \hline  
    \multirow{6}{*}{LLM}
      & \multirow{2}{*}{CLIP}   & mean & 37.96 & 39.06 & 13.31 & 25.16 \\
      &                         & max  & 40.58 & 47.40 & 14.61 & 26.98 \\
      & \multirow{2}{*}{DINO}   & mean & 31.83 & 20.55 & 6.24 & 21.93 \\
      &                         & max  & 33.02 & 26.41 & 8.17 & 20.89 \\
      & \multirow{2}{*}{DINOv2} & mean & 36.65 & 43.64 & \underline{16.74} & 21.27 \\
      &                         & max  & \textbf{44.83} & \textbf{54.31} & \textbf{17.52} & \textbf{28.55} \\
    \hline  
    \multirow{6}{*}{Naive}
      & \multirow{2}{*}{CLIP}   & mean & 34.68 & 42.32 & 11.08 & 25.76 \\
      &                         & max  & 39.65 & \underline{50.73} & 14.36 & 27.21 \\
      & \multirow{2}{*}{DINO}   & mean & 33.34 & 5.53 & 6.20 & 21.13 \\
      &                         & max  & 29.40 & 10.32 & 6.88 & 21.65 \\
      & \multirow{2}{*}{DINOv2} & mean & 35.35 & 40.75 & 15.05 & 23.24 \\
      &                         & max  & \underline{43.92} & 48.92 & 15.64 & \underline{28.22} \\
    \hline  
  \end{tabular}
  \label{tab:ablation1}
\end{table}
\vspace{-5pt}

\subsubsection{Different Components in S2C} 

As shown in Table~\ref{tab:ablation2}, we conduct a systematic evaluation of the proposed S2C on three change detection datasets. Based on the standard WSCD strategy, S2C introduces VFM-based spatial modeling capabilities for pseudo-label refinement, which significantly improves the binary change localization accuracy after pre-training. Although S2C cannot directly handle multi-class OVCD, it can be integrated into training-free frameworks to yield consistent performance improvements. These gains are mainly attributed to the effective suppression of false positives in change localization.

\begin{table}[ht]
  \centering
  \fontsize{9.3pt}{10pt}\selectfont
  \setlength{\tabcolsep}{4pt}
  \renewcommand{\arraystretch}{1.2}
  \caption{Ablation study on (mean) IoU of S2C and its integration with training-free OVCD models. 
  $\mathcal{L}$, $\mathcal{W}$, and $\mathcal{H}$ denote the LEVIR-CD, WHU-CD, and Hi-UCD mini datasets, respectively.}
  \label{tab:s2c_ablation}

  \begin{tabular}{c c|c c|c c c}
    \hline
    \multicolumn{2}{c|}{\multirow{2}{*}{OVCD Method}} &
      \multicolumn{2}{c|}{S2C Components} &
      \multirow{2}{*}{$\mathcal{L}$} & \multirow{2}{*}{$\mathcal{W}$} & \multirow{2}{*}{$\mathcal{H}$} \\
    \multicolumn{2}{c|}{} &
      \makebox[1.2cm][c]{Base} & \makebox[1.2cm][c]{Refine} & & & \\
    \hline
    \multicolumn{2}{c|}{—} & \checkmark & × & 43.26 & 61.84 & — \\
    \multicolumn{2}{c|}{—} & \checkmark & \checkmark & 44.67 & 64.83 & — \\
    \hline
    \multirow{3}{*}{M-C-I} &  & × & × & 36.63 & 40.60 & 13.94 \\
     & \textcolor{gray}{-W} & \checkmark & × & 40.74 & 62.44 & 14.79 \\
     & \textcolor{gray}{-W} & \checkmark & \checkmark & 41.48 & 64.43 & 15.85 \\
    \hline
    \multirow{3}{*}{OpenDPR} &   & × & × & 44.83 & 54.31 & 17.52 \\
     & \textcolor{gray}{-W} & \checkmark & × & \underline{47.07} & \underline{66.40} & \underline{17.74} \\
     & \textcolor{gray}{-W} & \checkmark & \checkmark & \textbf{48.13} & \textbf{68.28} & \textbf{19.69} \\
    \hline
  \end{tabular}
  \label{tab:ablation2}
\end{table}
\vspace{-5pt}

\subsubsection{Weak Supervision for Training-Free Methods}

As shown in Table~\ref{tab:ablation2}, we integrate S2C into M-C-I and OpenDPR in a plug-and-play manner, yielding their weakly supervised variants. For OpenDPR, introducing additional weak supervision brings significant performance improvements across all three datasets. In contrast, although the weakly supervised version of M-C-I also surpasses its original counterpart, it still performs worse than directly applying S2C for binary building change detection on LEVIR-CD and WHU-CD. This mainly stems from its limited capability in category identification, further validating our analysis in Section~\ref{sec:intro} that category identification errors have a more pronounced impact on overall OVCD performance than change localization errors.

\subsubsection{Impact of the Number of Prototypes}

\begin{figure}[ht]
	\centering
	\includegraphics[width=8.5cm, trim=0.1cm 0.1cm 0cm 0.1cm,clip]{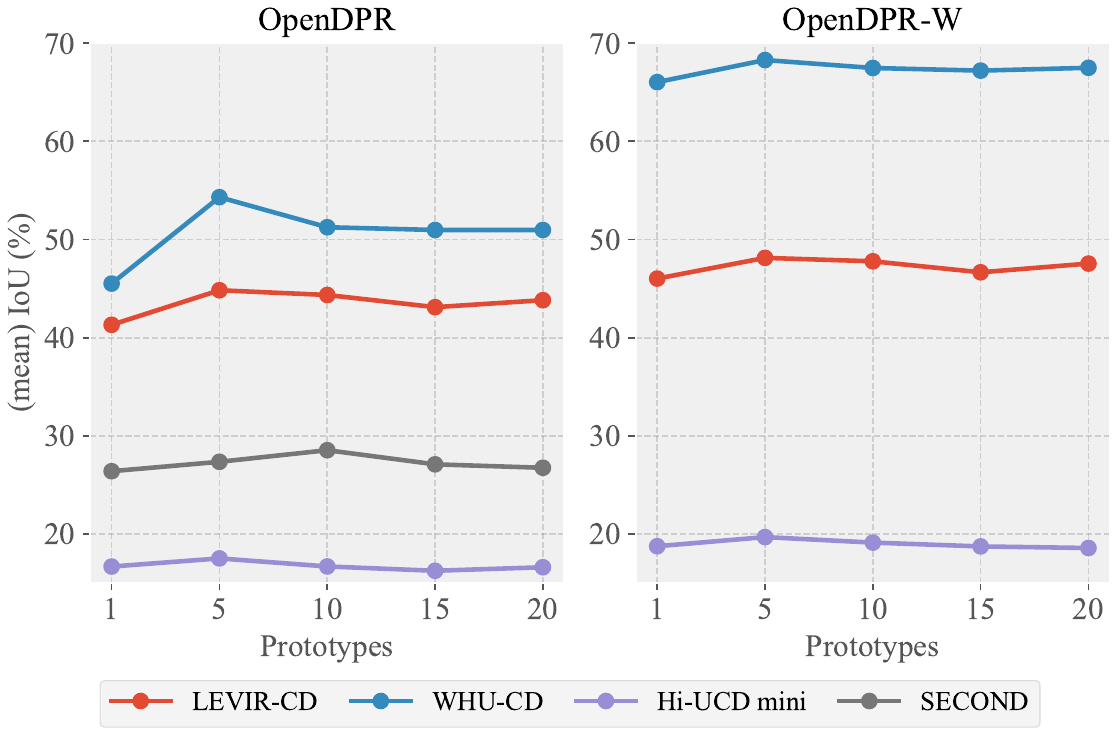}
	\caption{Prototype sensitivity analysis on OpenDPR and OpenDPR-W. Only OpenDPR is evaluated on SECOND.}
	\label{fig: prototype}
\end{figure}

To evaluate the impact of prototype quantity in prototype construction, we conduct a sensitivity analysis on the training-free OpenDPR and its weakly supervised variant OpenDPR-W across four datasets. As shown in Figure~\ref{fig: prototype}, both methods exhibit relatively stable performance trends under varying numbers of prototypes, with OpenDPR-W consistently achieving higher accuracy. Such robustness to the number of prototypes is particularly crucial in open-world scenarios, where the optimal quantity is often difficult to predefine.

\setlength{\abovecaptionskip}{4pt}  
\section{Conclusion}
\label{sec:Conclusions}

In this paper, we reformulate OVCD as a two-stage pipeline consisting of change localization and category identification. We reveal that category identification errors are the primary bottleneck, mainly due to the suboptimal performance of VLMs based on image-text matching in complex scenarios. To address this, we propose a training-free vision-centric diffusion-guided prototype retrieval framework named OpenDPR, which performs identification via similarity retrieval between change proposals and category prototypes in the visual space.
Moreover, the secondary bottleneck lies in inaccurate change localization, which arises from the absence of explicit change priors in VFMs. To mitigate this, we propose a spatial-to-change weakly supervised change detection module named S2C to adapt the spatial awareness of VFMs for change localization via pre-training. Integrating S2C into OpenDPR yields the weakly supervised variant OpenDPR-W, which further improves OVCD performance. Extensive experiments on four datasets validate the effectiveness of the proposed methods.

\section*{Acknowledgment}
This work was supported by the National Natural Science
Foundation of China under Grant No.42325105 and Grant No.42501475.

{
    \small
    \bibliographystyle{ieeenat_fullname}
    \bibliography{main}
}


\end{document}